%% file: arxiv.tex
\documentclass[11pt, a4paper, copyright, gdm]{google}

\usepackage[authoryear, sort&compress, round]{natbib}
\usepackage{amsmath,amssymb,mathtools}
\usepackage{graphicx}
\usepackage{booktabs}
\usepackage{placeins}
\usepackage{wrapfig}
\usepackage{microtype}
\usepackage{xspace}
\usepackage{hyperref}
\usepackage{url}

\newcommand{\method}{\textsc{Evo}\textsuperscript{2}\xspace}
\newcommand{\trainalgo}{\textsc{EvoTIDE}\xspace}
\newcommand{\trainalgorithm}{Tree-guided Interleaved Dual-role Evolution\xspace}
\newcommand{\applyedit}{\operatorname{Apply}}
\newcommand{\score}{\widehat{V}}
\newcommand{\policy}{\pi_{\theta}}

\newcommand{\qwen}{Qwen3.8-27B\xspace}
\newcommand{\polyglot}{Polyglot\xspace}
\newcommand{\swe}{SWE-bench Verified\xspace}
\newcommand{\lcb}{LiveCodeBench\xspace}
\newcommand{\lcm}{LoCoMo\xspace}
\newcommand{\lme}{LongMemEval\xspace}

\uselogo{} 

\title{\method: Evolving the Evolver through Weight-Harness Co-Evolution}

\renewcommand{\today}{2026-10-09}

\author[*,2]{Chejian Xu}
\author[1]{Shujian Zhang}
\author[1]{Wenxuan Zhou}
\author[1]{John Lambert}
\author[1]{Beck Chen}
\author[2]{Bo Li}
\author[1]{Andrew Hard}
\author[1]{Rajiv Mathews}

\affil[1]{\thepa{}{}}
\affil[2]{University of Illinois, Urbana-Champaign}
\affil[*]{Work done as a student researcher at \thepa}

\begin{abstract}
Self-evolving agents are increasingly used to improve language model systems
without human intervention through either weight updates or harness improvements.
Existing approaches typically pursue these directions separately: some methods
update model weights using self-generated experience, while others rewrite the
harness that defines how an agent uses prompts, tools, context, and execution
feedback. However, these approaches overlook the relationship between model
weights and the agent harness. To connect these processes and ensure effective and smooth agent self-evolving, we propose \method,
a framework for weight-harness co-evolution. For harness
updates, we design a unified dual-role pipeline that a single policy serves as both a Proposer,
which writes harness edits, and a Reflector, which diagnoses their effects and
guides revision. For weight updates, harness modifications and their execution
outcomes guide the training of both the Proposer and the Reflector.
We further develop \trainalgo, a training pipeline that recovers usable
trajectories through edit extraction and prefix continuation, then combines
SFT with interleaved GRPO across the two roles. Extensive experiments across
three coding benchmarks demonstrate that \method significantly outperforms
existing self-evolution methods. It achieves 82.1\% on \polyglot
and 48.3\% on \swe, compared with 62.6\% and 41.0\% for the
strongest baseline, respectively. We also show that the evolved harnesses
transfer to different tasks without further evolution and rank first in cross-benchmark
evaluations. Our dual-role harness improvement framework is also able to extend to
self-evolution with proprietary models, improving a GPT-5-mini agent's
performance from 24.4\% to 48.3\%.
In addition, \method generalizes to other tasks smoothly. We extend self-evolution to memory
tasks, achieving the best performance on the \lcm and \lme benchmarks.
These results demonstrate the applicability of the framework \method across domains
and its broad transferability across models and tasks.
\end{abstract}

\begin{document}

\begingroup
\renewcommand{\thefootnote}{\fnsymbol{footnote}}
\maketitle
\endgroup
\setcounter{footnote}{0}

\section{Introduction}
\label{sec:intro}


Self-evolving agents offer a way to improve language model systems through
their own experience and without human intervention. Existing approaches
typically follow two directions. The first updates \emph{model weights}:
a model generates solutions, selects useful outputs using task feedback,
and trains on them to improve its ability to solve similar problems
\citep{zelikman2022star,singh2024beyond}. The second modifies the
\emph{agent harness}: a model inspects and rewrites the program that defines
how an agent uses prompts, tools, context, and execution feedback
\citep{robeyns2025self,zhang2026darwin}. By evaluating these edits and retaining
useful changes, an agent improves its task performance through modifications
to its own code. These approaches accumulate experience in model parameters
and executable programs, respectively, but typically pursue the two forms
of improvement separately.

Treating these directions separately overlooks the relationship between model
weights and the agent harness. The model's capabilities shape the modifications
it proposes and how it interprets execution feedback. In turn, each evaluated
modification reveals which changes are useful, why an edit fails, and how
different improvements interact. This experience can guide the training of
the model responsible for evolution. In particular, proposing edits and
diagnosing their effects are skills that can themselves be learned from
earlier evolution attempts, connecting harness improvement to weight updates.

\begin{figure}[t]
    \centering
    \includegraphics[width=\linewidth]{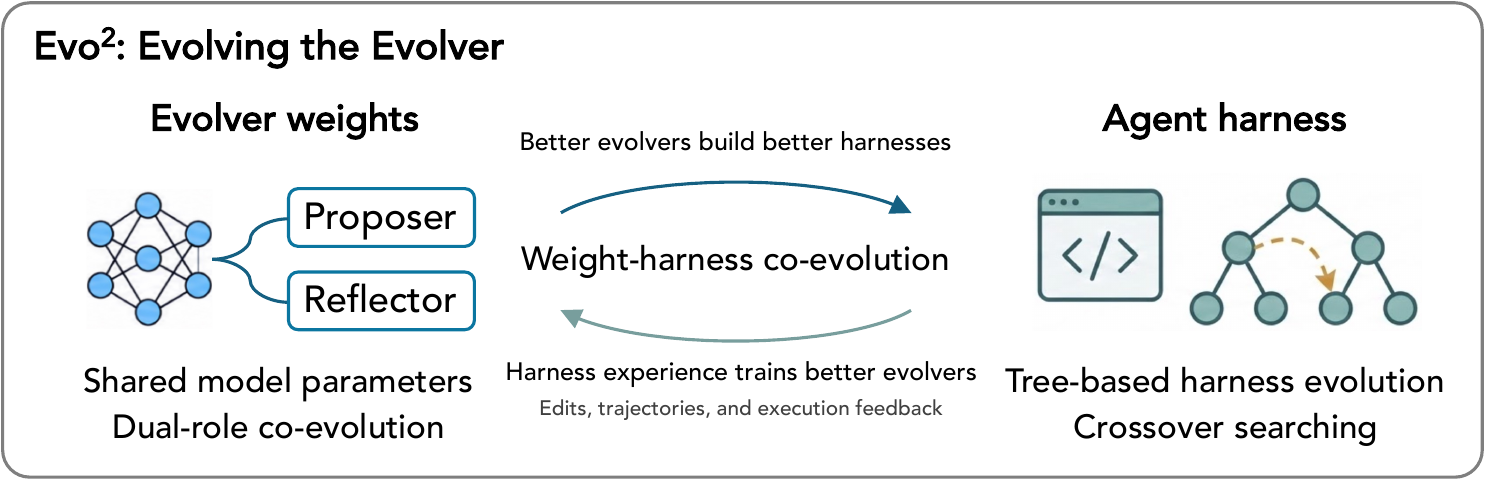}
    \caption{\textbf{Weight-harness co-evolution in \method.}
    A shared Proposer--Reflector policy evolves the agent harness. Harness edits
    and execution outcomes provide training experience for the policy, whose
    updated weights guide subsequent search. Crossover searching transfers
    useful strategies between branches. }
    \label{fig:coevolution}
\end{figure}

We introduce \method, a framework for \emph{weight-harness co-evolution}
in self-evolving agents (Figure~\ref{fig:coevolution}). We train the model
to become a better \emph{evolver}: to propose useful harness modifications,
interpret their outcomes, and revise them accordingly. A better evolver
builds better harnesses, whose modifications and execution outcomes provide
further experience for training the evolver. Harness evolution, therefore,
supplies training signals for weight updates, while the updated model guides
subsequent harness search.

For harness updates, \method uses a unified dual-role pipeline in which
a single policy serves as both a \emph{Proposer} and a \emph{Reflector}.
The Proposer writes harness edits, while the Reflector examines their
execution outcomes and provides feedback for further revision. Both roles
share parameters and learn from the effects of the modifications they produce.
We organize the search as a tree of harness variants and introduce
\emph{crossover searching} to combine useful strategies discovered on different
branches. The Reflector identifies a strategy that addresses a weakness in
the current harness, and the Proposer adapts it to that harness, allowing
each branch to build on discoveries made elsewhere in the tree.

For weight updates, we develop \trainalgo{}\ (\trainalgorithm) for weight updates, to train the
shared policy on experience collected during harness evolution. We collect
trajectories from both the Proposer and the Reflector, using edit extraction
and prefix continuation to recover usable trajectories from incomplete or
partially invalid rollouts. These trajectories support an SFT warm start,
followed by interleaved GRPO across the two roles. A maintained parent pool
exposes the policy to harnesses at different depths and performance levels,
so it learns both to improve strong candidates and to repair weaker ones.
The updated policy then generates the proposals and feedback used in
subsequent evolution rounds.

Extensive experiments across three coding benchmarks demonstrate that
\method outperforms existing self-evolution methods. It achieves 82.1\%
on \polyglot and 48.3\% on \swe, compared with 62.6\% and
41.0\% for the strongest respective baselines. The evolved harnesses also
transfer across benchmarks without further evolution. We further apply
the dual-role harness improvement framework to proprietary models with
fixed weights and extend self-evolution to memory tasks, achieving the
best performance on \lcm and \lme. These experiments demonstrate
both the applicability of the framework across domains and the
transferability of its evolved harnesses.

Our contributions are:
\begin{itemize}
    \item \textbf{Weight-harness co-evolution:} We introduce \method, which
    connects harness evolution with the training of a unified
    Proposer--Reflector policy through harness modifications and their
    execution outcomes.

    \item \textbf{Crossover searching:} We combine reflection and directed
    editing to adapt useful strategies across branches of the harness tree.

    \item \textbf{A training pipeline for the evolver:} We develop \trainalgo,
    which combines trajectory recovery, SFT, and interleaved GRPO with a
    parent pool spanning different depths and performance levels.

    \item \textbf{A reusable framework and evolved harnesses:} We demonstrate
    self-evolution on coding and memory tasks, harness transfer across
    benchmarks, and harness improvement with fixed proprietary models.
\end{itemize}

\section{Related Work}
\label{sec:related}

\paragraph{Self-improving language models.}
Self-improvement through training forms an iterative loop: a model generates
experience, learns from feedback, and uses its updated weights to collect the
next round of training data~\citep{gao2025survey,fang2025comprehensive}.
STaR and ReST$^{\mathrm{EM}}$
train on successful model outputs~\citep{zelikman2022star,singh2024beyond}, while
Absolute Zero and R-Zero develop tasks and solvers through
self-play~\citep{zhao2026absolute,huang2026r}. Other methods improve training
signals through trajectory selection, semantic voting, or learned
evaluators~\citep{zhongbetter,jiang2026semantic,wang2024self}.
EvolveR combines reusable experience with policy
reinforcement~\citep{wu2025evolver}. ReVeal, Agent0-VL, and Self-Distillation Zero
train generation together with verification or
revision~\citep{jin2026reveal,liu2025agent0,he2026self}; EvoLM and CoNL co-train
generation and feedback mechanisms within shared
policies~\citep{li2026evolm,sui2026conversation}. \method applies shared-role
training to harness modification: \trainalgo uses supervised fine-tuning and
interleaved GRPO to train a unified Proposer--Reflector model on evaluated edits
and their outcomes.

\paragraph{Self-evolving agent harnesses.}
Agents can also accumulate improvements in the harness that manages their
prompts, tools, and control flow. They propose modifications, evaluate the
resulting agent, and use the outcomes to guide further changes.
SICA edits its own implementation~\citep{robeyns2025self}, DGM maintains an
archive for branching self-modification~\citep{zhang2026darwin}, and HGM selects
promising parents using estimated descendant
performance~\citep{wang2026huxley}. Meta-Harness searches over code using prior
candidates and execution traces~\citep{leemeta}. GEPA combines reflection with
prompt evolution~\citep{agrawal2026gepa}, while GEA shares experience across
evolving agents~\citep{weng2026group}. The Red Queen G\"odel Machine extends
evolution to changing evaluation utilities~\citep{iacob2026red}. Prior work also
demonstrates harness transfer across tasks and
models~\citep{wang2026huxley,leemeta,weng2026group}. \method makes harness search
a source of self-improvement for the evolver itself: crossover searching lets the
Reflector select a donor strategy and the Proposer adapt it to a recipient
harness, while the resulting experience trains both roles to guide subsequent
search. We evaluate the
framework's applicability across coding and memory domains as well as the transfer of
its evolved harnesses across benchmarks and executors.

\section{Method}
\label{sec:method}

\subsection{Overview}
\label{sec:method:overview}

\begin{figure}[t]
    \centering
    \includegraphics[width=\linewidth]{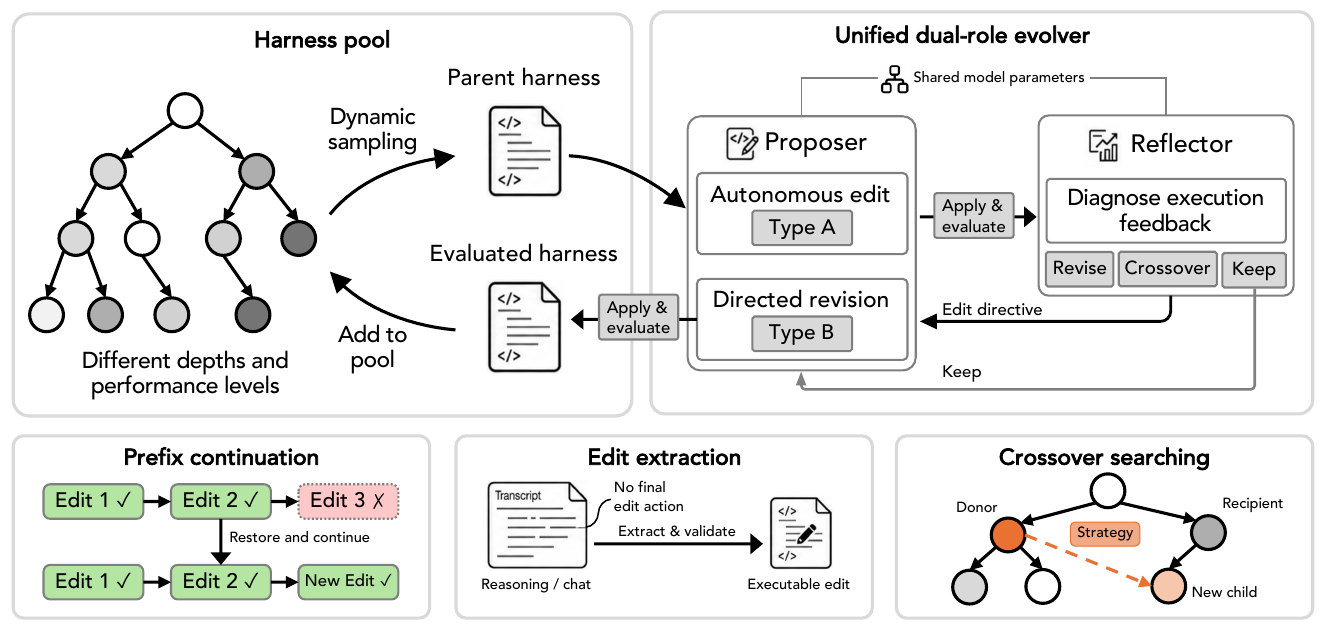}
    \caption{\textbf{Harness evolution and trajectory collection.}
    Parents sampled from the harness pool receive autonomous Type~A edits from
    the Proposer and are evaluated with a fixed task executor. The Reflector
    uses execution feedback to direct revision or crossover, or to keep the candidate without further editing. Both roles
    share model parameters, and evaluated candidates extend the pool. The lower
    panels illustrate prefix continuation, edit extraction, and crossover
    searching. }
    \label{fig:method-pipeline}
\end{figure}

\method integrates harness evolution with weight evolution
(Figure~\ref{fig:method-pipeline}). A harness, $h$, is the
executable program that manages an agent's prompts, tools, context, and control
flow. A language model policy, $\policy$, serves as the evolver, acting as both
Proposer and Reflector through different role prompts. We train its parameters,
$\theta$, on the outcomes of harness modifications. The task executor, denoted
$M_{\mathrm{exec}}$, runs inside each candidate harness and remains fixed during
co-evolution, so candidate performance measures the effect of changing the
harness.

Harness evolution proceeds through iterative tree search. Each node stores a
harness and its task evaluation records; an edge records the edits made to its
parent. At each iteration, we sample a parent, use the Proposer to modify it, and
evaluate the resulting candidate. The Reflector then examines the execution
feedback and recommends a revision, integration of a strategy from another
branch, or retention of the candidate. The Proposer implements any recommended
change, which is evaluated in turn. The resulting harnesses extend the tree,
while the role trajectories and task outcomes provide data for weight evolution.
\trainalgo uses these data for supervised fine-tuning followed by interleaved
reinforcement learning. Updated weights are used in subsequent rounds of search
and data collection.

\subsection{Harness evolution through dual-role self-modification}
\label{sec:method:harness}

\paragraph{Proposer.}
The Proposer operates in two modes. In \emph{Type A}, it receives a parent harness
$h_p$ and its measured task performance, chooses a modification, and produces
edits $e_0$ to obtain an initial candidate $h_0$. In \emph{Type B}, it receives
a harness and an explicit edit directive $d$ from the Reflector. It implements
that directive against the
current source through edits $e_1$, producing a revised candidate $h_1$.
Type A develops the ability to identify improvements independently; Type B
develops the ability to turn feedback into working code. Both modes can generate
a sequence of file edits. 

\paragraph{Reflector.}
The Reflector receives the candidate source, the edits relative to its parent,
the change in task performance, and execution traces from relevant successes
and failures. It identifies the strategy attempted by the Proposer, diagnoses
remaining problems, and chooses one of three actions: \textsc{revise} provides a
directive to improve the candidate; \textsc{crossover} provides a directive to
incorporate a strategy from another branch; and \textsc{keep} retains the candidate
without another edit. A revision directive describes a concrete change and the
execution evidence that motivates it. Its quality is measured by the performance
of the harness produced when the Proposer follows it. This also allows the
Reflector to learn when an additional modification is unlikely to help.

Let $z_p$ contain the parent's evaluation records and $z_0$ contain the candidate's
edit diff and execution feedback. With available donor strategies $\mathcal{S}$,
the shared policy implements the cycle
\begin{equation}
    \begin{aligned}
        e_0 &\sim \policy(\cdot\mid P_A,h_p,z_p),
        & h_0 &= \applyedit(h_p,e_0),\\
        (m,d,j) &\sim \policy(\cdot\mid R,h_p,h_0,z_0,\mathcal{S}),
        &&\\
        e_1 &\sim \policy(\cdot\mid P_B,h_0,d),
        & h_1 &= \applyedit(h_0,e_1).
    \end{aligned}
    \label{eq:dual-role-cycle}
\end{equation}
Here $P_A$, $P_B$, and $R$ specify role prompts; all calls use the same model
parameters $\theta$. We write the Reflector's decision compactly as its action $m$, directive
$d$, and donor identifier $j$, which is used only for crossover. The last line is
executed for revision or crossover; for \textsc{keep}, $h_1=h_0$ and no further
edit is sampled. During training, counterpart calls use the frozen snapshot
described in Section~\ref{sec:method:training}.

\paragraph{Crossover searching.}
Different branches may discover complementary improvements. To make these
discoveries available to the Reflector, we maintain a library
$\mathcal{S}=\{s_j\}$ of strategies extracted from evaluated harnesses. Each entry
describes a mechanism and its
observed effect, together with whether it is already present in the current
candidate. Given this context, the Reflector can select a donor strategy that
addresses a problem revealed by the candidate's execution. Its directive $d$
specifies how to integrate the selected $s_j$ into $h_0$. The Type B Proposer
then adapts the mechanism to the candidate's source and combines it with the
existing behavior. We evaluate the combined harness to determine whether the
integration helps. A crossover child retains the recipient as its tree parent
and records the donor strategy, so it can receive further revision or crossover.

\subsection{Weight evolution through \texorpdfstring{\trainalgo}{Evo2-TIDE}}
\label{sec:method:weight}

\trainalgo turns the trajectories generated during harness evolution into
training data for the shared policy. We first collect an initial dataset and
perform supervised fine-tuning (SFT). We then continue collecting experience from
the evolving tree while training both roles with group relative policy
optimization (GRPO)~\citep{shao2024deepseekmath}.

\subsubsection{Data collection}
\label{sec:method:data}

For each evolution attempt, we record the role prompts, model responses, applied
edits, parent and child harnesses, and task evaluation results. One
propose-evaluate-reflect-revise cycle provides an autonomous Proposer
trajectory, a Reflector trajectory, and, when a revision is requested, a directed
Proposer trajectory. The same evaluated harness can also serve as a parent in a
later round or contribute a strategy to the crossover library.

\paragraph{Recovering Proposer trajectories.}
An otherwise useful rollout may fail to produce an executable edit because of
how the response is expressed. We use two recovery procedures during data
collection. \emph{Edit extraction} handles rollouts that end without an explicit
edit action: we search the recorded reasoning and chat messages for concrete
file edits and convert them into the required action format. \emph{Prefix
continuation} handles a sequence of edits in which a later edit is invalid, for
example because of a malformed tool call or an incorrect edit format. We truncate
the rollout after the last successfully applied edit, restore the corresponding
intermediate harness, and resume generation from that prefix. This preserves the
valid earlier edits while allowing the model to regenerate the failed suffix.
Recovered proposals pass the same application, startup, and task evaluation
checks as other candidates before entering the training data.

\paragraph{Collecting Reflector trajectories.}
Evaluated proposals become reflection cases, including edits that improve
performance, leave it unchanged, or cause regressions. For each case, we generate
feedback and evaluate the candidate obtained by implementing it; a \textsc{keep}
response retains the initial candidate. These outcomes provide supervision for
the Reflector's diagnosis and choice of action. Connecting strategies from other
branches adds crossover cases, so the dataset includes both local revisions and
the integration of previously discovered mechanisms.

\subsubsection{Reward design}
\label{sec:method:rewards}

Both roles are rewarded by the change in task performance caused by their
decisions. Let $\score(h)\in[0,1]$ be the mean externally evaluated task score of harness
$h$; for tasks with a binary correctness signal, this is its success rate.
For a valid Proposer edit from $h$ to $h'$, and a valid Reflector intervention
from $h_0$ to $h_1$, the execution rewards are
\begin{equation}
    r_P^{\mathrm{exec}}=\score(h')-\score(h),
    \qquad
    r_R^{\mathrm{exec}}=\score(h_1)-\score(h_0).
    \label{eq:execution-rewards}
\end{equation}
For Type A, the Proposer is compared with its sampled parent; for Type B, it is
compared with the candidate it was instructed to revise. A \textsc{keep} action
has $h_1=h_0$ and receives zero execution reward. We evaluate each comparison on
the same tasks under matched execution conditions, averaging repeated runs when
available.

Malformed outputs and edits that cannot be applied or started receive a negative
reward. No-op edits, unusable harnesses, and invalid crossover directives are
also penalized. Type B receives a small additional bonus or penalty according to
whether the edit touches a file named in the directive. For Reflector
interventions, positive gains are credited only when the traces contain evidence
that the added mechanism ran; otherwise the gain is set to zero. 

\subsubsection{Training process}
\label{sec:method:training}

\paragraph{Supervised fine-tuning.}
We initialize the shared policy with a mixture of successful autonomous edits,
directed edits, and Reflector responses selected using their downstream
evaluation results. The model is trained to predict the recorded response from
its role prompt and harness context. We balance the training mixture across
roles and modes to establish both the editing and reflection behavior needed for
online training.

\paragraph{Tree maintenance and parent sampling.}
During online training, evaluated candidates extend the harness tree and refresh
a bounded parent pool. We retain nodes at multiple depths so the model encounters
both early modifications and harnesses with several accumulated changes. Within
each depth, we reserve part of the pool for the strongest candidates and sample
the remainder from other executable candidates. This retains weaker harnesses
that need repair as well as strong harnesses that can be improved further. Let
$\mathcal{P}_r$ be the retained pool at round $r$, and let
$\mathcal{N}_r=\mathcal{P}_r\setminus\{h_{\mathrm{root}}\}$ contain its non-root
harnesses. The root has a permanent place and an allocated sampling share
$\rho\in(0,1)$. With weights
$w(h)=\epsilon_p+|\score(h)-\tfrac{1}{2}|$, where $\epsilon_p>0$, parent samples
follow
\begin{equation}
    p_r(h)=
    \begin{cases}
        \rho, & h=h_{\mathrm{root}},\\[2pt]
        (1-\rho)\dfrac{w(h)}{\sum_{u\in\mathcal{N}_r}w(u)},
        & h\in\mathcal{N}_r.
    \end{cases}
    \label{eq:parent-sampling}
\end{equation}
If $\mathcal{N}_r$ is empty, we sample only the root. This distribution gives
both high and low-scoring parents more exposure while keeping every retained
parent available. Updating the pool across rounds lets the training distribution
develop with the harnesses produced by the model.

\paragraph{Interleaved role training.}
We optimize a single policy for both roles. At the beginning of training round
$r$, we set $\bar{\theta}_r=\theta_r$ and freeze $\pi_{\bar{\theta}_r}$ to provide
the counterpart behavior needed to evaluate training samples. For Proposer
training, this copy provides Reflector directives for Type B prompts. For
Reflector training, it acts as the
Proposer that implements sampled directives. Each optimization step uses a
Proposer group and a Reflector group, with advantages computed separately for
each prompt. Both role updates change the live parameters $\theta$, while
$\bar{\theta}_r$ stays fixed. After the round yields $\theta_{r+1}$, we refresh
the snapshot by setting $\bar{\theta}_{r+1}=\theta_{r+1}$ and collect new cases.
An improved Proposer therefore generates new modifications for the Reflector to
learn from, while an improved Reflector
generates new directives for the Proposer. Both skills improve through successive
rounds of this interaction.

\paragraph{Balancing modes within each role.}
We interleave Type A and Type B Proposer prompts so that training covers both
autonomous improvement and implementation of feedback. For the Reflector, we
sample half of the inputs with an empty strategy set and half with crossover
strategies. We also construct paired SFT prompts: when a target response is valid
in either context, the same response is used with and without the strategy set.
This teaches the model to retain a useful diagnosis even when donor information
is unavailable. Targets that depend on a specific crossover donor are used only
with the corresponding strategy context. The mixture includes revision,
crossover, and retention decisions, and is refreshed as new trajectories become
available.

\paragraph{Policy optimization.}
For a role prompt $q$, we sample a group of $G$ responses $y_1,\ldots,y_G$ from
the current policy and compute rewards $r_1,\ldots,r_G$. We use the group mean baseline
$A_i=r_i-\frac{1}{G}\sum_{j=1}^{G}r_j$ and normalize the token losses by a fixed
maximum response length $L_{\max}$. We average the losses from the
two role groups and take one update per sampled batch. Truncated responses are
masked from the loss. 
All responses within a group share the same role
and input, so their advantages compare alternative edits or reflections for the
same harness context.

The trained evolver can run the same search procedure on a new initial harness
given an evaluator for its tasks. An evolved harness can also be deployed
directly on new tasks with compatible model and tool interfaces. These two forms correspond to the outputs of co-evolution: weights that encode how to
improve a harness, and harnesses that incorporate the improvements
found during search.

\section{Experiments}
\label{sec:experiments}

\subsection{Experimental Settings}
\label{sec:exp-setup}

\textbf{Models.}
Our main experiments use \qwen, served in FP8 with greedy decoding, as the
fixed task executor. We use the same model as the evolver: the untrained base
model for the baselines and the model fine-tuned by \trainalgo for \method,
where a single policy serves as both Proposer and Reflector. For memory tasks,
we additionally use a Qwen3-Embedding-0.6B retriever. In the proprietary-model
experiments, GPT-5-mini and Claude Haiku 4.5 each serve as both executor and
evolver, with their weights fixed throughout evolution.

\textbf{Benchmarks and metrics.}
We evaluate coding harness evolution on \polyglot, \swe (SWE), and \lcb (LCB).
We report held-out pass@1 on \polyglot tasks over three evaluation runs,
the resolved rate on \swe instances, and pass@1 on LCB problems.
Proprietary models are evaluated on a 60-task \swe subset. For memory tasks,
we evolve harnesses on \lcm and report mean F1 on three held-out conversations. We also evaluate the same harnesses on \lme{}-S
using accuracy over 500 instances, with GPT judging and three evaluation runs.

\textbf{Baselines.}
We compare against four representative self-evolving agent frameworks:
\textbf{SICA}~\citep{robeyns2025self}, which edits its own implementation along
a single lineage; \textbf{DGM}~\citep{zhang2026darwin}, which maintains an
open-ended archive and branches self-modifications from stored agents;
\textbf{HGM}~\citep{wang2026huxley}, which selects parents based on the
estimated performance of their descendants; and
\textbf{GEA}~\citep{weng2026group}, which enables concurrently evolving agents
to share experience. All baselines use the untrained model as the evolver and
follow their own selection rules.

\input{tabs/table1_self_evolution}

\subsection{Self-Evolution on Coding Benchmarks}
\label{sec:exp-coding}

\textbf{\method evolves the strongest harness on every benchmark.}
Table~\ref{tab:self-evolution} reports the held-out performance of the best
harness evolved by each method on each benchmark.
On \polyglot, \method achieves 82.1\% pass@1, compared with 62.6\% for the
strongest baseline, GEA. On \swe, it resolves 48.3\% of held-out instances,
compared with 41.0\% for HGM. On \lcb, \method reaches 58.0\% pass@1, while
the four baselines range from 39.0\% to 56.0\%. These results demonstrate
consistent improvements across all three coding benchmarks.


\input{tabs/table2_benchmark_transfer}

\subsection{Harness Transfer across Benchmarks}
\label{sec:exp-transfer}

For each method and source benchmark, we take the harness reported in
Table~\ref{tab:self-evolution} and evaluate it on the other two benchmarks.
Table~\ref{tab:benchmark-transfer} reports all six transfer directions.

\textbf{\method achieves the best transfer performance in five of six directions.}
The harness evolved on \swe achieves 72.4 on \polyglot and 53.0 on LCB,
while the harness evolved on LCB achieves 74.0 on \polyglot and 43.0 on
\swe. In each case, \method outperforms every baseline, with margins ranging
from 2.5 to 18 points. These results show that the improvements discovered
during evolution remain useful beyond the tasks used to select the harness.


\begin{table}[t]
    \centering
    \small
    \setlength{\tabcolsep}{3pt}
    \begin{minipage}[t]{0.48\textwidth}
        \vspace{0pt}
        \caption{Self-evolution with proprietary models on a 60-task SWE-bench
        Verified subset. We report the resolved rate on each model. In \method, each model serves as executor,
        Proposer, and Reflector with fixed weights.} 
        \label{tab:proprietary}
        \begin{tabular*}{\linewidth}{@{\extracolsep{\fill}}lcc@{}}
            \toprule
            Method & GPT-5-mini & \shortstack{Claude\\Haiku 4.5} \\
            \midrule
            SICA & 24.4 & 42.2 \\
            DGM  & 23.9 & 43.3 \\
            HGM  & 24.3 & 45.0 \\
            GEA  & 19.4 & 44.4 \\
            \midrule
            \method (ours) & \textbf{48.3} & \textbf{47.8} \\
            \bottomrule
        \end{tabular*}
    \end{minipage}\hfill
    \begin{minipage}[t]{0.48\textwidth}
        \vspace{0pt}
        \caption{Self-evolution on memory tasks. \lcm: held-out F1 of the
        harness evolved and selected on \lcm. \lme: accuracy of the same
        harness on \lme-S without further evolution or selection. }
        \label{tab:memory}
        \begin{tabular*}{\linewidth}{@{\extracolsep{\fill}}lcc@{}}
            \toprule
            Method & \shortstack{\lcm{}\\F1 $\uparrow$}
                & \shortstack{\lme{}\\Accuracy $\uparrow$} \\
            \midrule
            SICA & 54.6 & 59.7 \\
            DGM  & 55.8 & 60.3 \\
            HGM  & 56.2 & 64.4 \\
            GEA  & 53.2 & 48.9 \\
            \midrule
            \method (ours) & \textbf{60.4} & \textbf{71.5} \\
            \bottomrule
        \end{tabular*}
    \end{minipage}
\end{table}

\subsection{Self-Evolution with Proprietary Models}
\label{sec:exp-proprietary}

Table~\ref{tab:proprietary} reports self-evolution results on the 60-task
\swe subset with GPT-5-mini and Claude Haiku 4.5. In each column, the named
model performs both task execution and harness evolution. For \method, it
serves as executor, Proposer, and Reflector with fixed weights; the dual-role
reflection and crossover search therefore constitute the differences from
the baselines. We conduct three independent searches per method and model on
a small search subset and report the mean over seeds. Each baseline evaluates
its top-ranked harness, while \method reports the better result of its two
top-ranked harnesses.

\textbf{Dual-role search improves proprietary agents with fixed weights.}
With GPT-5-mini, \method achieves a 48.3\% resolved rate, compared with
19.4--24.4\% for the baselines, nearly doubling the strongest baseline's
performance. With Claude Haiku 4.5, whose unevolved harness already performs
strongly, \method reaches 47.8\%, compared with 42.2--45.0\% for the baselines.
These results demonstrate that harness search provides improvements on its
own. When the evolver can be trained, weight evolution provides additional
gains, as shown in Section~\ref{sec:exp-analysis}.

\subsection{Self-Evolution on Memory Tasks}
\label{sec:exp-memory}

To evaluate the framework in another domain, we instantiate it with a memory
harness and perform self-evolution on \lcm. We retain the Proposer--Reflector
interaction, crossover search, and training procedure, while adapting the
editable harness and task evaluator to memory tasks. Each search evaluates
1{,}500 questions drawn from seven training conversations. Table~\ref{tab:memory} reports the results.

\textbf{\method outperforms the strongest baseline on \lcm and transfers better to \lme.}
On \lcm, \method achieves 60.4 F1, compared with 56.2 for HGM and 53.2--55.8
for SICA, DGM, and GEA. On \lme, which is unseen during both evolution and
selection, every \method search achieves a score between 67.9 and 73.9,
exceeding the HGM mean of 64.4. GEA performs below the other baselines.
All numbers are percentages averaged over independent searches.

\input{tabs/table6_weight_ablation}

\subsection{In-Depth Analysis of \method}
\label{sec:exp-analysis}

\begin{wrapfigure}{r}{0.43\textwidth}
    \centering
    \includegraphics[width=\linewidth]{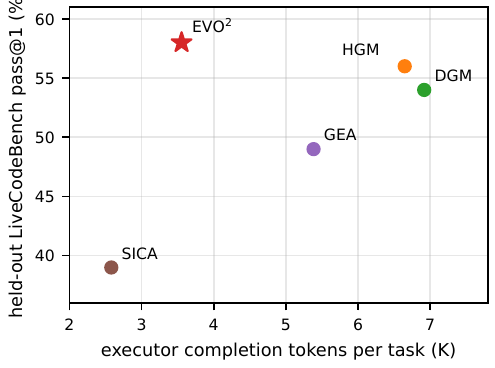}
    \caption{\textbf{Held-out accuracy versus inference cost.}
    Held-out \lcb pass@1 of the harnesses in Table~\ref{tab:self-evolution},
    plotted against mean executor completion tokens per held-out task.
    \method achieves the highest accuracy, while HGM and DGM use about
    $1.9\times$ as many tokens and $2.6\times$ and $3.6\times$ as many LLM
    calls per task, respectively.}
    \label{fig:efficiency}
\end{wrapfigure}

\textbf{Weight evolution improves the evolver on every benchmark.}
Table~\ref{tab:weight-ablation} isolates the contribution of \trainalgo.
Each comparison fixes the search procedure, the executor, and the
parent harnesses, varying only whether the evolver's
weights are trained. The trained evolver produces stronger champions on
\polyglot ($+16.3$ points) and \lcm ($+9.3$), while
its \lcm champion transfers better to \lme ($+10.6$). Crossover search
provides a further 4.1-point improvement on \polyglot in
Table~\ref{tab:self-evolution}. The trained evolver
produces better children on average.

\textbf{Performance gains do not require additional inference compute.}
Figure~\ref{fig:efficiency} compares held-out \lcb accuracy with the
inference cost of each evolved harness. The \method harness uses 3.6K
completion tokens and 10.2 LLM calls per task.
HGM and DGM achieve lower accuracy despite using
6.6K--6.9K tokens and 26--37 calls per task. The same pattern holds on
\polyglot, where our champion uses fewer calls and tokens than the unevolved
harness: 36 versus 43 calls and 3.6K versus 4.0K tokens per task. These results
show that harness evolution improves how effectively the agent uses a fixed
inference budget.







\FloatBarrier
\section{Conclusion}
\label{sec:conclusion}

We presented \method, a framework for weight-harness co-evolution in
self-evolving agents. A single policy serves as both Proposer and Reflector,
learning from harness modifications and their execution outcomes to improve
its ability to propose edits and guide revision. Crossover searching combines
useful strategies across branches, while \trainalgo trains the shared policy
through trajectory recovery, SFT, and interleaved GRPO. Experiments on coding
and memory tasks demonstrate improved task performance and transfer of evolved
harnesses across benchmarks. The dual-role search also supports harness
improvement with fixed proprietary models. Our analyses show that weight
evolution improves search reliability and that crossover searching reduces
the evaluation budget needed to reach strong performance. Evolving the evolver
allows experience to accumulate in both the executable harness and the policy
that guides future evolution.

\bibliography{iclr2027_conference}

\end{document}

%% file: tabs/table1_self_evolution.tex
\begin{table}[tbp]
    \centering
    \small
    \caption{Self-evolution on coding benchmarks. Each column reports the
    held-out performance of the harness evolved on that benchmark, using \qwen.
    All scores are percentages; higher is better.} 
    \label{tab:self-evolution}
    \begin{tabular*}{\linewidth}{@{\extracolsep{\fill}}lccc@{}}
        \toprule
        Method & \shortstack{\polyglot{}\\Pass@1 $\uparrow$}
               & \shortstack{\swe{}\\Resolved $\uparrow$}
               & \shortstack{\lcb{}\\Pass@1 $\uparrow$} \\
        \midrule
        SICA & 60.2 & 20.5 & 39.0 \\
        DGM  & 60.2 & 30.0 & 54.0 \\
        HGM  & 61.8 & 41.0 & 56.0 \\
        GEA  & 62.6 & 25.0 & 49.0 \\
        \midrule
        \method (ours) & \textbf{82.1} & \textbf{48.3} & \textbf{58.0} \\
        \bottomrule
    \end{tabular*}
\end{table}

%% file: tabs/table2_benchmark_transfer.tex
\begin{table}[tbp]
    \centering
    \small
    \setlength{\tabcolsep}{4pt}
    \caption{Transfer across coding benchmarks, using \qwen. The upper header identifies
    the benchmark used to evolve and select the best harness; the lower header
    identifies its evaluation benchmark. Each method contributes one harness
    per source, evaluated on both other benchmarks without further evolution.
    Scores use the target benchmark's metric from Table~\ref{tab:self-evolution}.}
    \label{tab:benchmark-transfer}
    \begin{tabular*}{\linewidth}{@{\extracolsep{\fill}}lcccccc@{}}
        \toprule
        & \multicolumn{2}{c}{From \polyglot}
        & \multicolumn{2}{c}{From SWE-BenchVerif.}
        & \multicolumn{2}{c}{From LCB} \\
        \cmidrule(lr){2-3}\cmidrule(lr){4-5}\cmidrule(l){6-7}
        Method & SWE & LCB & \polyglot & LCB & \polyglot & SWE \\
        \midrule
        SICA & 19.0 & 33.0 & 52.0 & 49.0 & 43.9 & 29.0 \\
        DGM  & 20.0 & \textbf{45.0} & 69.9 & 45.0 & 63.4 & 19.0 \\
        HGM  & 20.0 & 41.0 & 53.7 & 50.0 & 63.4 & 16.0 \\
        GEA  & 16.0 & \textbf{45.0} & 48.8 & 42.0 & 61.0 & 25.0 \\
        \midrule
        \method (ours) & \textbf{21.0} & 44.0 & \textbf{72.4} & \textbf{53.0} & \textbf{74.0} & \textbf{43.0} \\
        \bottomrule
    \end{tabular*}
\end{table}

%% file: tabs/table6_weight_ablation.tex
\begin{table}[t]
    \centering
    \small
    \setlength{\tabcolsep}{4pt}
    \caption{Effect of weight evolution. Each row compares the untrained evolver
    (the base \qwen acting as Proposer and Reflector) with the evolver
    trained by \trainalgo.}
    \label{tab:weight-ablation}
    \begin{tabular*}{\linewidth}{@{\extracolsep{\fill}}lccc@{}}
        \toprule
        Benchmark & Untrained & Trained & $\Delta$ ($P$) \\
        \midrule
        \polyglot & 61.8 & \textbf{78.0} & $+16.3$ ($.001$) \\
        \swe & 46.0 & \textbf{51.0} & $+5.0$ ($.086$) \\
        \lcm & 47.2 & \textbf{56.5} & $+9.3$ \\
        \lme & 57.1 & \textbf{67.7} & $+10.6$ \\
        \bottomrule
    \end{tabular*}
\end{table}